\documentclass[final]{cvpr}

\usepackage{times}
\usepackage{epsfig}
\usepackage{graphicx}
\usepackage{amsmath}
\usepackage{amssymb}

\usepackage{caption}
\usepackage{enumitem}
\usepackage{xspace}
\usepackage{subfigure}
\usepackage{hhline}
\usepackage[dvipsnames]{xcolor}
\usepackage{balance}

\usepackage[toc,page]{appendix}

\newcommand{\posa}{{\mbox{POSA}}\xspace}
\newcommand{\posaAcronym}{{\color{black}``Pose with prOximitieS and contActs''}}
\newcommand{\modelname}{\posa}
\newcommand{\modelnameAcronym}{\posaAcronym}

\newcommand{\websiteURL}{\mbox{\url{https://posa.is.tue.mpg.de}}}

\newcommand{\code}{\xspace{code}\xspace}

\newcommand{\jargonContact}{{{contact}}\xspace}

\newcommand{\hsi}{{{HSI}}\xspace}
\newcommand{\hoi}{{{HOI}}\xspace}

\newcommand{\pytorch}{\xspace{\mbox{PyTorch}}\xspace}

\newcommand{\agora}{\xspace{\mbox{AGORA}}\xspace}
\newcommand{\vposer}{\xspace{\mbox{VPoser}}\xspace}
\newcommand{\renderpeople}{\xspace{\mbox{Renderpeople}}\xspace}

\newcommand{\scenegrok}{\xspace{\mbox{SceneGrok}}\xspace}

\newcommand{\smplX}{\xspace{\mbox{SMPL-X}}\xspace}
\newcommand{\smplifyX}{{\mbox{SMPLify-X}}\xspace}
\newcommand{\mano}{{\mbox{MANO}}\xspace}

\newcommand{\video}{{{{video}}}\xspace}
\newcommand{\supmat}{{\mbox{{Sup.~Mat.}}}\xspace}

\newcommand{\twoD}{{2D}\xspace}
\newcommand{\threeD}{{3D}\xspace}

\newcommand{\mColl}{\mathcal{P}}

\newcommand{\rgb}{RGB\xspace}
\newcommand{\rgbD}{\mbox{RGB-D}\xspace}

\newcommand{\mesh}{M}
\newcommand{\scene}{s}

\newcommand{\body}{b}
\newcommand{\face}{f}
\newcommand{\hand}{h}
\newcommand{\imgRGB}{I}

\newcommand{\trans}{{\tau}}

\usepackage{esvect}
\newcommand{\feat}{f}
\newcommand{\featGen}{f_\text{Gen}}
\newcommand{\featMap}{{feature map}\xspace}
\newcommand{\featMaps}{{feature maps}\xspace}

\newcommand{\featCon}{{f_c}}
\newcommand{\featDst}{{f_d}}

\newcommand{\featSem}{f_s}

\newcommand{\closestSceneP}{{P_\scene}}
\newcommand{\meshScene}{{\mesh_\scene}\xspace}
\newcommand{\meshBody}{{\mesh_\body}\xspace}
\newcommand{\surfaceScene}{{\mathcal{S}_\scene}\xspace}
\newcommand{\surfaceBody}{{\mathcal{S}_\body}\xspace}

\newcommand{\bodyV}{V_\body}
\newcommand{\queryBodyV}{V_\body^{{i}}}

\newcommand{\ncomps}{{{12}}}
\newcommand{\betasNumb}{{10}}

\newcommand{\place}{{\mbox{PLACE}}\xspace}

\newcommand{\prox}{{\mbox{PROX}}\xspace}
\newcommand{\proxe}{{\mbox{PROX-E}}\xspace}

\newcommand{\pigraphs}{{\mbox{PiGraphs}}\xspace}
\newcommand{\bps}{{BPS}\xspace}

\renewcommand{\etal}{et al.\xspace}
\renewcommand{\ie}{i.e.\xspace}
\renewcommand{\eg}{e.g.\xspace}
\renewcommand{\wrt}{w.r.t.\xspace}

\newcommand{\redify}[1]{\textcolor{black}{{#1}}}

\usepackage[pagebackref=true,breaklinks=true,colorlinks,bookmarks=false]{hyperref}

\def\cvprPaperID{5950} %
\def\confYear{CVPR 2021}
\begin{document}
\title{Populating 3D Scenes by Learning Human-Scene Interaction}

\author{Mohamed Hassan \quad Partha Ghosh \quad Joachim Tesch \quad Dimitrios Tzionas \quad Michael J. Black\\
Max Planck Institute for Intelligent Systems, T{\"u}bingen, Germany\\
{\tt\small \{mhassan, pghosh, jtesch, dtzionas, black\}@tuebingen.mpg.de}
}

\twocolumn[
{
    \renewcommand\twocolumn[1][]{#1}
    \maketitle
    \centering
    \begin{minipage}{1.00\textwidth}
    \centering
        \includegraphics[trim=000mm 000mm 000mm 000mm, clip=false, width=1.00 \textwidth]{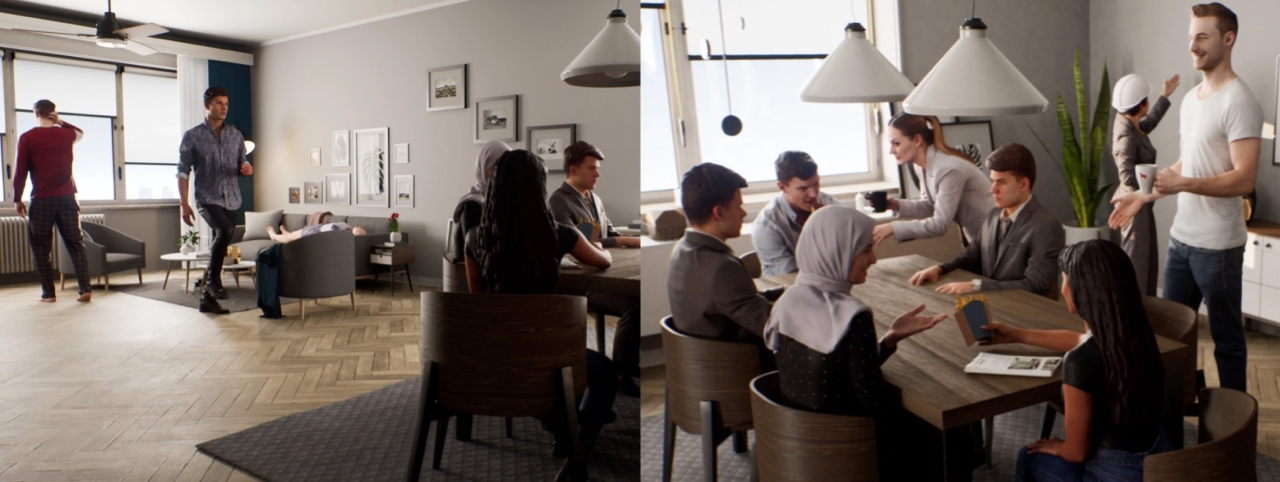}
    \end{minipage}
 \vspace{-0.1in}
   \captionof{figure}{
				\modelname automatically places \threeD people in \threeD scenes such that the interactions between the people and the scene are both geometrically and semantically correct.  
				\modelname exploits a new learned representation of human bodies that explicitly models how bodies interact with scenes. 
    }\label{fig:teaser}
    \vspace*{+02.00em}
}
]

\maketitle
\thispagestyle{empty}
\pagestyle{empty}

\begin{abstract}
Humans live within a \threeD space and constantly interact with it to perform tasks. 
Such interactions involve physical contact between surfaces that is semantically meaningful. 
Our goal is to learn how humans interact with scenes and leverage this to enable virtual characters to do the same.
To that end, we introduce a novel \mbox{Human-Scene} Interaction (\hsi) model that encodes proximal relationships, called \modelname for \modelnameAcronym. 
The representation of interaction is \mbox{body-centric}, which enables it to generalize to new scenes.
Specifically, \modelname augments the \smplX parametric human body model such that, for every mesh vertex, it encodes
(a)	the contact probability with the scene surface and
(b)	the corresponding semantic scene label.
We learn \modelname with a VAE conditioned on the \smplX vertices, and train on the \prox dataset, which contains \smplX meshes of people interacting with \threeD scenes, and the corresponding scene semantics from the \proxe dataset.
We demonstrate the value of \modelname with two applications. 
First, we automatically place \threeD scans of people in scenes. 
We use a \smplX model fit to the scan as a proxy and then find its most likely placement in \threeD.
\modelname provides an effective representation to search for ``affordances'' in the scene that match the likely \jargonContact relationships for that pose. 
We perform a perceptual study that shows significant improvement over the state of the art on this task.
Second, we show that \modelname's learned representation of body-scene interaction supports monocular human pose estimation that is consistent with a \threeD scene, improving on the state of the art.
Our model and \code are available for research purposes at \websiteURL. 
\end{abstract}

\section{Introduction}
Humans constantly interact with the world around them. 
We move by walking on the ground; we sleep lying on a bed; we rest sitting on a chair; we work using touchscreens and keyboards. 
Our bodies have evolved to exploit the affordances of the natural environment and we design objects to better ``afford'' our bodies. 
While obvious, it is worth stating that these physical interactions involve \jargonContact. 
Despite the importance of such interactions, existing representations of the human body do not explicitly represent, support, or capture them.

In computer vision, human pose is typically estimated in isolation from the \threeD scene, while in computer graphics \threeD scenes are often scanned and reconstructed without people.
Both the recovery of humans in scenes and the automated synthesis of realistic people in scenes remain challenging problems.
Automation of this latter case would reduce animation costs and open up new applications in augmented reality.
Here we take a step towards automating the realistic placement of \threeD people in \threeD scenes with realistic contact and semantic interactions (Fig.~\ref{fig:teaser}).
We develop a novel \emph{body-centric} approach that relates \threeD body shape and pose to possible world interactions.
Learned parametric \threeD human models \cite{Anguelov05,joo2018total,SMPL:2015,SMPL-X:2019} represent the shape and pose of people accurately. 
We employ the \smplX \cite{SMPL-X:2019} model, which includes the hands and face, as it supports reasoning about contact between the body and the world.

While such body models are powerful,  we make three \emph{key observations}. 
First, 	human models like \smplX \cite{SMPL-X:2019} do not explicitly model \jargonContact. 
Second,	not all parts of the body surface are equally likely to be in contact with the scene.
Third, 	the poses of our body and scene semantics are highly intertwined.
		Imagine a person sitting on a chair; 			body contact likely includes the buttocks, probably also the back, and maybe the arms. 
		Think of someone opening a door;		their feet are likely in \jargonContact with the floor, and their hand is in \jargonContact with the doorknob. 

Based on these observations, we formulate a novel model, that makes human-scene interaction (\hsi) an explicit and integral part of the body model.
The key idea is to encode \hsi in an ego-centric representation built in \smplX.
This effectively extends the \smplX model to capture \jargonContact and the semantics of \hsi in a body-centric representation.
We call this \modelname for \emph{\modelnameAcronym}.
Specifically, for every vertex on the body and every pose, \modelname defines a probabilistic \featMap that encodes the probability that the vertex is in contact with the world and the distribution of semantic labels associated with that contact.

\modelname is a conditional Variational \mbox{Auto-Encoder} (cVAE), conditioned on \smplX vertex positions.
We train on the \prox dataset \cite{PROX:2019}, which contains $20$ subjects, fit with  \smplX meshes, interacting with $12$ real \threeD scenes. 
We also train \modelname using use the scene semantic annotations provided by the \proxe dataset \cite{zhang2020generating}.
Once trained, given a posed body, we can sample likely contacts and semantic labels for all vertices.
We show the value of  this representation with two challenging applications. 

First, we focus on automatic scene population as illustrated in Fig.~\ref{fig:teaser}.
That is, given a \threeD scene and a body in a particular pose, where in the scene is this pose most likely?
As demonstrated in Fig.~\ref{fig:teaser} we use  \smplX bodies fit to commercial 3D scans of people \cite{agora2020}, and then, conditioned on the body, our cVAE generates a target \modelname~\featMap. 
We then search over possible human placements while minimizing the discrepancy between the observed and target \featMaps.
We quantitatively compare our approach to  PLACE \cite{PLACE:3DV:2020}, which is SOTA on a similar task,
and find that \posa has higher perceptual realism.

Second, we use \modelname for monocular \threeD human pose estimation in a \threeD scene.
We build on the \prox method \cite{PROX:2019} that \mbox{hand-codes} \jargonContact points, and replace these with our learned \featMap, which functions as an \hsi prior. 
This automates a heuristic process, while producing lower pose estimation errors than the original \prox method.

To summarize, \modelname is a novel model that intertwines \smplX pose and scene semantics with \jargonContact. 
To the best of our knowledge, this is the first learned human body model that incorporates \hsi in the model.
We think this is important because such a model can be used in all the same ways that models like \smplX are used but now with the addition of body-scene interaction.
The key novelty is posing \hsi as part of the body representation itself.
Like the original learned body models, \modelname provides a platform that people can build on.
To facilitate this, our model and \code are available for research purposes at \websiteURL. 

\section{Related Work}

\textbf{Humans \& Scenes in Isolation:}
For \emph{scenes} \cite{zollhofer2018sotaRconstructionRGBD}, most work focuses on their \threeD shape in isolation, \eg 
on rooms devoid of humans \cite{armeni2017joint,Matterport3D,dai2017scannet,replica19arxiv,xiazamirhe2018gibsonenv} or 
on objects that are not grasped \cite{calli2015benchmarking,chang2015shapenet,choi2016large,xiang2016objectnet3d}. 
For \emph{humans}, there is extensive work on 
capturing 	\cite{dfaust:CVPR:2017,ionescupapavaetal2014,AMASS:2019,sigal_ijcv_10b} or 
estimating 	\cite{Review_Moeslund_2006,Sarafianos:Survey:2016}
their \threeD shape and pose, but outside the context of scenes. 

\textbf{Human Models:}
Most work represents \threeD humans as body skeletons \cite{ionescupapavaetal2014,sigal_ijcv_10b}. 
However, the \threeD body surface is important for physical interactions. 
This is addressed by learned parametric \threeD body models \cite{Anguelov05,joo2018total,SMPL:2015,STAR:ECCV:2020,SMPL-X:2019,Xu_2020_CVPR}. 
For interacting humans we employ \smplX \cite{SMPL-X:2019}, which models the body with full face and hand articulation. 

\textbf{\hsi Geometric Models}:
We focus on the spatial relationship between a human body and objects it interacts with. 
Gleicher        \cite{gleicher1998retargetting} 	uses contact constraints for early work on motion \mbox{re-targeting}. 
Lee		  \etal \cite{lee2006motionPatches}     	generate novel scenes and motion by deformably stitching ``motion patches'', comprised of scene patches and the skeletal motion in them. 
Lin       \etal \cite{lin2012sketching}         	generate \threeD skeletons sitting on \threeD chairs, by manually drawing \twoD skeletons and fitting \threeD skeletons that satisfy collision and balance constraints. 
Kim       \etal \cite{shape2pose2014}           	automate this, by detecting sparse contacts on a \threeD object mesh and fitting a \threeD skeleton to contacts while avoiding penetrations. 
Kang      \etal \cite{kang2014environment}      	reason about the physical comfort and environmental support of a \threeD humanoid, through force equilibrium. 
Leimer    \etal \cite{leimer2020pose}           	reason about pressure, frictional forces and body torques, to generate a \threeD object mesh that comfortably supports a given posed \threeD body. 
Zheng     \etal \cite{zheng2015ergonomics}      	map high-level ergonomic rules to low-level contact constraints and deform an object to fit a \threeD human skeleton for force equilibrium. 
Bar-Aviv  \etal \cite{bar2006functional}        	and Liu \etal \cite{liu2015indirect} use an interacting agent to describe object shapes through detected contacts \cite{bar2006functional}, or relative  distance and orientation metrics \cite{liu2015indirect}. 
Gupta     \etal \cite{gupta2011workspace} 	   	estimate human poses ``afforded'' in a depicted room, by predicting a \threeD scene occupancy grid, and computing support and penetration of a \threeD skeleton in it. 
Grabner   \etal \cite{grabner2011chair}         	detect the places on a \threeD scene mesh where a \threeD human mesh can sit, modeling interaction likelihood with GMMs and proximity and intersection metrics. 
Zhu       \etal \cite{zhu2016inferringForces}   	use FEM simulations for a \threeD humanoid, to learn to estimate forces, and reasons about sitting comfort. 

Several methods focus on dynamic interactions \cite{alAsqhar2013relationship,ho2010spatial,pirk2017understanding}. 
Ho        \etal \cite{ho2010spatial} 		   	compute an ``interaction mesh'' per frame, through Delaunay tetrahedralization on human joints and unoccluded object vertices; 
											   	minimizing their Laplacian deformation maintains spatial relationships. 
											   	Others follow an object-centric approach \cite{alAsqhar2013relationship,pirk2017understanding}. 
Al-Asqhar \etal \cite{alAsqhar2013relationship} 	sample fixed points on a \threeD scene, using proximity and ``coverage'' metrics, 
											   	and encode \threeD human joints as transformations \wrt~these. 
Pirk      \etal \cite{pirk2017understanding}    	build functional object descriptors, by placing ``sensors'' around and on \threeD objects, that ``sense'' \threeD flow of particles on an agent. 

\textbf{\hsi Data-driven Models}:
Recent work takes a data-driven approach. 
Jiang     \etal \cite{jiang2013hallucinated}	   	learn to estimate human poses and object affordances from an \rgbD scene, for \threeD scene label estimation. 
\scenegrok 		\cite{scenegrok2014savva} 		learns action-specific classifiers to detect the likely scene places that ``afford'' a given action. 
Fisher 	  \etal \cite{fisher2015activity} 		use \scenegrok and interaction annotations on CAD objects, to embed noisy \threeD room scans to CAD mesh configurations. 
\pigraphs 		\cite{savva2016pigraphs} 		maps pairs of \{verb-object\} labels to ``interaction snapshots'', \ie~ \threeD interaction layouts of objects and a human skeleton. 
Chen 	  \etal \cite{chen2019holistic++} 		map \rgb images to ``interaction snapshots'', using Markov Chain Monte Carlo with simulated annealing to optimize their layout. 
iMapper 			\cite{iMapper2018} 				maps \rgb videos to dynamic ``interaction snapshots'', 
												by learning ``scenelets'' on \pigraphs data and fitting them to videos. 
Phosa \cite{zhang2020phosa} infers spatial arrangements of humans and objects from a single image.
Cao 		  \etal \cite{caoHMP2020} 				map an \rgb scene and \twoD pose history to \threeD skeletal motion, by training on video-game data. 
Li        \etal \cite{Li2019puttingHumansScenes}	follow \cite{Wang2017binge} to collect \threeD human skeletons consistent with \twoD/\threeD scenes of \cite{SUNCG:2017,Wang2017binge}, 
												and learn to predict them from a color and/or depth image.
Corona    \etal \cite{corona2020contextAware}    use a graph attention model to predict motion for objects and a human skeleton, and their evolving spatial relationships.

Another \hsi variant is Hand-Object Interaction (\hoi); 
we discuss only recent work \cite{Brahmbhatt_2020_ECCV,corona2020ganhand,GRAB:2020,wang2019guibasHOI}. 
Brahmbhatt 	\etal 	\cite{Brahmbhatt_2020_ECCV}	capture fixed \threeD hand-object grasps, and learn to predict contact; 
												features based on 
												object-to-hand mesh distances outperform skeleton-based variants. 
												For grasp generation, 2-stage networks are popular \cite{Mousavian2019graspnet}.
Taheri 		\etal 	\cite{GRAB:2020}				capture moving \smplX \cite{SMPL-X:2019} humans grasping objects, and predict \mano \cite{MANO:2017} hand grasps for object meshes, whose \threeD shape is encoded with \bps \cite{prokudin2019efficient}.
Corona 		\etal	\cite{corona2020ganhand}		generate \mano grasping given an object-only \rgb image; 
												they first predict the object shape and rough hand pose (grasp type), and then they refine the latter with contact constraints \cite{hasson19_obman} and an adversarial prior. 

Closer to us, PSI \cite{zhang2020generating} and PLACE \cite{PLACE:3DV:2020} populate \threeD scenes with \smplX \cite{SMPL-X:2019} humans. 
Zhang 		\etal 	\cite{zhang2020generating} 	train a cVAE to estimate humans from a depth image and scene semantics. 
												Their model provides an implicit encoding of \hsi. 
Zhang 		\etal 	\cite{PLACE:3DV:2020}, 		on the other hand, explicitly encode the scene shape and human-scene proximal relations with \bps \cite{prokudin2019efficient}, but do not use semantics. 
Our key difference to \cite{PLACE:3DV:2020,zhang2020generating} is our human-centric formulation; inherently this is more portable to new scenes. 
Moreover, instead of the sparse \bps distances of \cite{PLACE:3DV:2020}, we use dense body-to-scene \jargonContact, and also exploit scene semantics like \cite{zhang2020generating}. 

\section{Method}

\subsection{Human Pose and Scene Representation}		\label{sec:method_Represent_Human_Scene}		Our training data corpus is a set of $n$ pairs of \threeD meshes 
\begin{equation}
 \mathcal{M} = \{ \{M_{b,1}, M_{s,1}\}, \{M_{b,2}, M_{s,2}\},\ldots, \{M_{b,n}, M_{s,n}\} \} \nonumber
\end{equation}
comprising body meshes 
$M_{b,i}$ and scene meshes $M_{s,i}$. 
We drop the index, $i$, for simplicity when we discuss meshes in general.
These meshes approximate human body surfaces $\surfaceBody$ and scene surfaces $\surfaceScene$. 
Scene meshes $\meshScene = (V_{\scene}, F_{\scene}, L_{\scene})$ have a varying number of vertices $N_{\scene} = |V_{\scene}|$ 
and triangle connectivity $F_{\scene}$ to model arbitrary scenes.
They also have  per-vertex semantic labels $L_{\scene}$. 
Human meshes are represented by \smplX \cite{SMPL-X:2019}, \ie a differentiable function 
$M(\theta, \beta, \psi) : \mathbb{R}^{|\theta| \times |\beta| \times |\psi|} \,\to\, \mathbb{R}^{(N_\body \times 3)}$ 
parameterized by pose $\theta$, shape $\beta$ and facial expressions $\psi$. 
The pose vector $\theta=(\theta_\body, \theta_\face, \theta_{lh}, \theta_{rh})$ is comprised of body, $\theta_\body \in \mathbb{R}^{66}$, and face parameters, $\theta_\face \in \mathbb{R}^{9}$, in \mbox{axis-angle} representation, 
while $\theta_{lh}, \theta_{rh} \in \mathbb{R}^\ncomps$  parameterize the poses of the left and right hands respectively in a low-dimensional pose space. 
The shape parameters, $\beta \in \mathbb{R}^{\betasNumb}$, represent coefficients in a \mbox{low-dimensional} shape space learned from a large corpus of human body scans. 
The joints, $J(\beta)$, of the body in the canonical pose are regressed from the body shape.
The skeleton has $55$ joints, consisting of $22$ body joints, $30$ hand joints, and $3$ head joints (neck and eyes). 
The mesh is posed using this skeleton and linear blend skinning. 
Body meshes $\meshBody = (\bodyV, F_{\body})$ have a fixed topology with $N_{\body} = |\bodyV| = 10475$ vertices $\bodyV \in \mathbb{R}^{(N_{\body} \times 3)}$ and triangles $F_{\body}$, \ie all human meshes are in correspondence.
For more details, we refer the reader to \cite{SMPL-X:2019}. 

\subsection{\modelname Representation for \hsi}		\label{sec:method_Represent__Interaction}	We encode the relationship between the human mesh $\meshBody = (\bodyV, F_{\body})$ and the scene mesh $\meshScene = (V_{\scene},
F_{\scene}, L_{\scene})$ in an egocentric \featMap $f$ that encodes per-vertex features on the \smplX mesh $\meshBody$. 
We define $f$ as:
\begin{align}
\feat &:\left(\bodyV, \meshScene \right) \rightarrow \left[ \featCon, \featSem\right], \label{eq:feat_2}
\end{align}
where $\featCon$ is the contact label and $\featSem$ is the semantic label of the contact point. $N_f$ is the feature dimension.

For each vertex $i$ on the body, $\queryBodyV$,  we find its closest scene point 
$\closestSceneP = \operatorname {argmin}_{\closestSceneP   \in \surfaceScene}   \| \closestSceneP - \queryBodyV \|$. 
Then we compute the distance $\featDst$:
\begin{equation}
	\featDst  =  \| \closestSceneP - \queryBodyV \|	\in \mathbb{R}. \label{eq:dist}
\end{equation}
Given $\featDst$, we can compute whether a $\queryBodyV$ is in contact with the scene or not, with $\featCon$: 
\begin{equation}
\featCon = 
\begin{cases} 
1 & \featDst \leq \text{\it Contact Threshold}, \\
0 &  \featDst >\text{\it Contact Threshold} .
\end{cases}
\end{equation}
The contact threshold is chosen empirically to be $5$ cm.
The semantic label of the contacted surface $\featSem$ is a one-hot encoding of the object class:
\begin{equation}
	\featSem = \left \{  0,1\right \}^{N_o}, \label{eq:sem}
\end{equation}
where $N_o$ is the number of object classes.
The sizes of $\featCon$, $\featSem$, and $f$ are $1$, $40$ and $41$ respectively.
All the features are computed once offline to speed training up.
A visualization of the proposed representation is in Fig.~\ref{fig:rep}.

\newcommand{\posaReprHeight}{0.430}
\begin{figure}[t]
	\centering
	\includegraphics[ height=\posaReprHeight \linewidth]{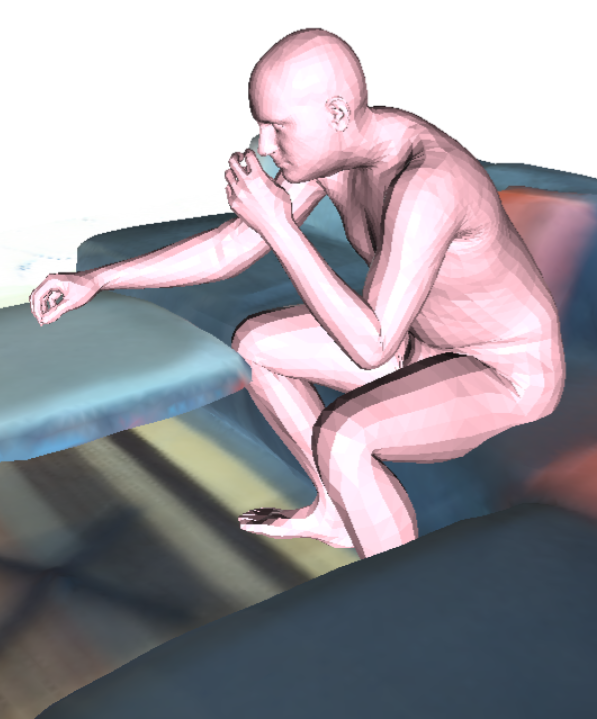}
	\includegraphics[ height=\posaReprHeight \linewidth]{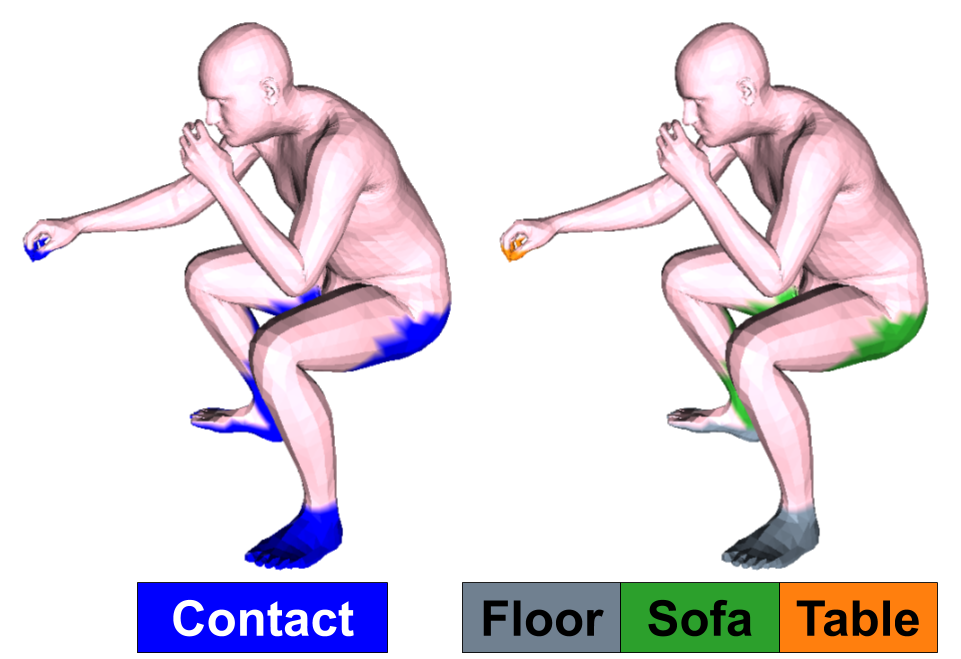}
\vspace{-0.24in}
	\caption{
		Illustration of our proposed representation. 
		From left to right: An example of a \smplX mesh $\meshBody$ in a scene $\meshScene$, contact $\featCon$, and scene semantics $\featSem$ on it. 
		For $\featCon$, blue means the body vertex is likely in contact. 
		For $\featSem$, the colors correspond to the scene semantics.
	}
	\label{fig:rep}
\end{figure}

\subsection{Learning}								\label{sec:method_Learning}					Our goal is to learn a probabilistic function from body pose and shape to the feature space of contact and semantics.
That is, given a body, we want to sample labelings of the vertices corresponding to likely scene contacts and their corresponding semantic label.
Note that this function, once learned, only takes the body as input and not a scene -- it is a {\it body-centric} representation.

To train this we use the \prox \cite{PROX:2019} dataset, which contains bodies in \threeD scenes. 
We also use the scene semantic annotations from the \proxe dataset \cite{zhang2020generating}. For each body mesh $\meshBody$, we factor out the global translation and rotation $R_y$ and $R_z$ around the $y$ and $z$ axes.  The rotation $R_x$ around the $x$ axis is essential for the model to differentiate between, \eg, standing up and lying down. 

Given pose and shape parameters in a given training frame, we compute a $\meshBody = M(\theta,\beta,\psi)$.  
This gives vertices $\bodyV$ from which we compute the \featMap that encodes whether each $\queryBodyV$ is in contact with the scene or not, and the semantic label of the  scene contact point $\closestSceneP$. 

We train a conditional Variational Autoencoder (cVAE), where we condition the \featMap on the vertex positions, $\bodyV$, which are a function of the body pose and shape parameters.
Training optimizes the encoder and decoder parameters to minimize $\mathcal{L}_{\mathrm{total}}$ using gradient descent:
\begin{align}
	\mathcal{L}_{\mathrm{total}}							&=  \alpha*\mathcal{L}_{KL} + \mathcal{L}_{rec}, 											\\
	\mathcal{L}_{KL} 									&=  KL(Q(z|\feat, \bodyV) || p(z)), 														\\
	\mathcal{L}_{rec}\left( \feat, \hat{\feat}\right)		&=  \lambda_c * \sum_i \text{BCE} \left(\featCon^i, \hat{\featCon}^i \right) \nonumber 	\\
														& + \lambda_s * \sum_i \text{CCE} \left(\featSem^i, \hat{\featSem}^i \right), \label{eq:l_recon}
\end{align}
where $\hat{\featCon}$ and $\hat{\featSem}$ are the reconstructed contact and semantic labels, {\it KL} denotes the Kullback Leibler divergence, and $\mathcal{L}_{rec}$ denotes the reconstruction loss.
BCE and CCE are the binary and categorical cross entropies respectively.
The $\alpha$ is a hyperparameter inspired by Gaussian $\beta$-VAEs \cite{higgins2017beta}, which regularizes the solution; here $\alpha=0.05$.
$\mathcal{L}_{rec}$ encourages the reconstructed samples to resemble the input, while $\mathcal{L}_{KL}$ encourages $Q(z|\feat,\bodyV)$ to match a prior distribution over $z$, which  is Gaussian in our case. We set the values of $\lambda_c$ and $\lambda_s$ to $1$.

\begin{figure}[t]
	\centering
	\includegraphics[trim=000mm 000mm 000mm 000mm, clip=true, width=1.00 \linewidth]{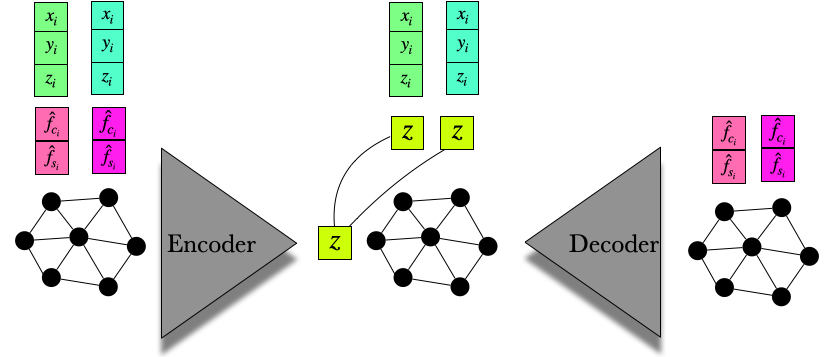}	
	\caption{
				cVAE architecture. 
				For each vertex on the body mesh, we concatenate the vertex positions $x_i, y_i, z_i$, the contact label $\featCon$, and the corresponding semantic scene label $\featSem$. 
                The latent vector $z$ is concatenated to the vertex positions, and the result passes to the decoder which reconstructs the input features $\hat{\featCon}$, $\hat{\featSem}$.
	}
	\label{fig:cVAE_arch}
\end{figure}

Since $f$ is defined on the vertices of the body mesh $\meshBody$, this enables us to use graph convolution as our building block for our VAE.
Specifically, we use the Spiral Convolution formulation introduced in \cite{bouritsas2019neural, gong2019spiralnet++}. 
The spiral convolution operator for node $i$ in the body mesh is defined as:
\begin{equation}
	\feat_k^i = \gamma_k \left( \|_{j \in S(i,l)} \feat_{k-1}^j \right),
\end{equation}
where $\gamma_k$ denotes layer $k$ in a multi-layer perceptron (MLP) network,  and $\|$ is a concatenation operation of the features of neighboring nodes, $S(i,l)$. 
The spiral sequence $S(i,l)$ is an ordered set of $l$ vertices around the central vertex $i$. 
Our architecture is shown in Fig.~\ref{fig:cVAE_arch}. More implementation details are included in the \supmat For details on selecting and ordering vertices, please see \cite{gong2019spiralnet++}.

\section{Experiments}

We perform several experiments to investigate the effectiveness and usefulness of our proposed representation and model under different
use cases, namely generating \hsi features, automatically placing 3D people in scenes,
and improving monocular \threeD human pose estimation.

\subsection{Random Sampling}							\label{sec:random_samples}			
\begin{figure}
	\centering	
	\includegraphics[width=0.99 \linewidth]{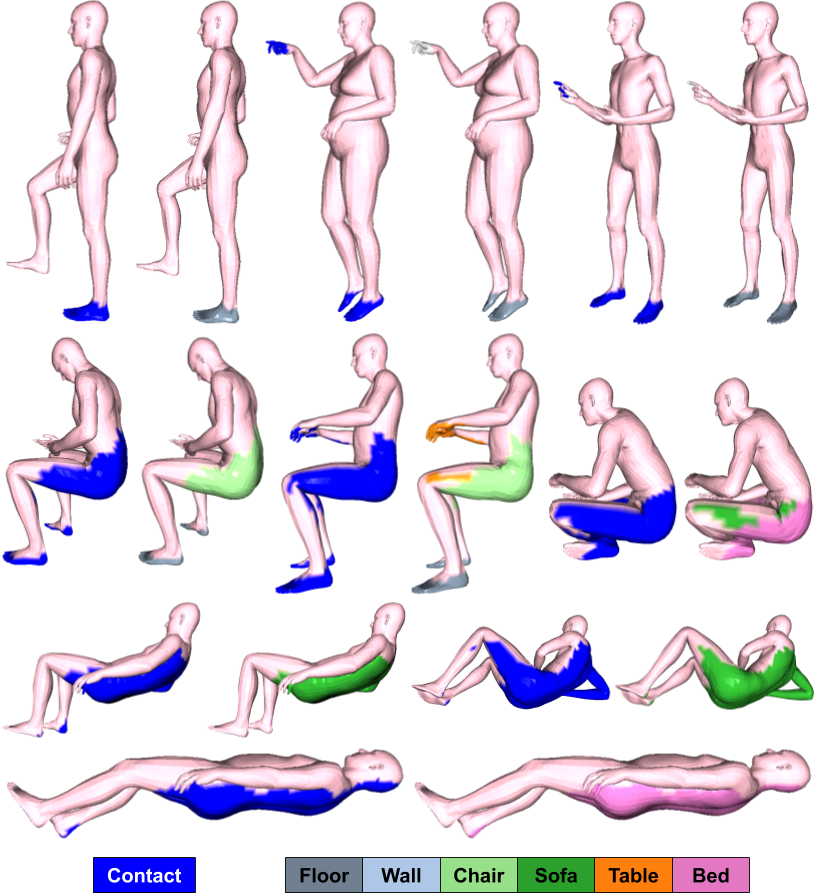}	
	\caption{
		Random samples from our trained cVAE. 
		For each example (image pair) we show from left to right: $\featCon$ and $\featSem$.
		The color code is at the bottom.  
		For $\featCon$, blue means contact, while pink means no contact.  
		For $\featSem$, each scene category has a different color.
	}
	\label{fig:random_samples}
\end{figure}

We evaluate the generative power of our model by sampling different feature maps conditioned on novel poses using our trained decoder $P(\featGen|z,\bodyV)$, where $z \sim \mathcal{N}(0,I) $ and $\featGen$ is the randomly generated \featMap.
This is equivalent to answering the question: 
``In this given pose, which vertices on the body are likely to be in contact with the scene, and what object would they contact?'' 
Randomly generated samples are shown in Fig.~\ref{fig:random_samples}. 

We observe that our model generalizes well to various poses.
For example, notice that when a person is standing with one hand pointing forward, our model predicts the feet and the hand to be in contact with the scene. 
It also predicts the feet are in \jargonContact with the floor and hand is in \jargonContact with the wall. 
However this changes for the examples when a person is in a lying pose. 
In this case, most of the vertices from the back of the body are predicted to be in \jargonContact (blue color) with a bed (light purple) or a sofa (dark green).

These features are predicted from the body alone; there is no notion of ``action'' here.  Pose alone is a powerful predictor of interaction.
Since the model is probabilistic, we can sample many possible feature maps for a given pose.

\subsection{Affordances: Putting People in Scenes}					\label{sec:affordance_detection}		
Given a posed 3D body and a 3D scene, can we place the body in the scene so that the pose makes sense in the context of the scene?
That is does the pose match the affordances of the scene \cite{grabner2011chair,shape2pose2014,kjellstrom2011visual}?
Specifically, given a scene, $\meshScene$, semantic labels of objects present, and a body mesh, $\meshBody$, our method finds where in $\meshScene$ this given pose is likely to happen. 
We solve this problem in two steps. 

First, given the posed body, we use the decoder of our cVAE to generate a \featMap by sampling $P(\featGen|z,\bodyV)$ as in Sec.~\ref{sec:random_samples}. 
Second, we optimize the objective function:
\begin{equation}
E(\trans, \theta_0, \theta) =  \mathcal{L}_{\mathit{afford}} + \mathcal{L}_{\mathit{pen}} + \mathcal{L}_{\mathit{reg}}, \label{eq:afford}
\end{equation}
where $\trans$ is the body translation, $\theta_0$ is the global body orientation and $\theta$ is the body pose.
The afforance loss $\mathcal{L}_{\mathit{afford}}=$ %
\begin{equation}
\lambda_1 * ||{\featGen}_{c} \cdot \feat_d||_2^2 + \lambda_2 * \sum_i \text{CCE} \left({\featGen}_{s}^{i}, \feat_s^i \right), \label{eg:afford_loss} 
\end{equation}
$\feat_d$        and $\feat_s$ 			are the observed distance and semantic labels, which are computed using Eq.~\ref{eq:dist} and Eq.~\ref{eq:sem} respectively. 
${\featGen}_{c}$ and ${\featGen}_{s}$ 	are the generated contact and semantic labels, and $\cdot$ denotes dot product.
$\lambda_1$ and $\lambda_2$ are $1$ and $0.01$ respectively.
$\mathcal{L}_{pen}$ 						is a penetration penalty to discourage the body from penetrating the scene:
\begin{equation}
	\mathcal{L}_{\mathit{pen}} = \lambda_{\mathit{pen}} * \sum_{\feat_d^i < 0} \left(\feat_d^i\right)^2. \label{eg:penetration}
\end{equation}
$\lambda_{pen} = 10$. $\mathcal{L}_{reg}$ is a regularizer  that encourages the estimated pose to remain close to the initial pose $\theta_{\text{init}}$ of $\meshBody$:
\begin{equation}
\mathcal{L}_{\mathit{reg}} = \lambda_{\mathit{reg}} * ||\theta - \theta_{\text{init}}||_2^2.
\end{equation}
Although the body pose is given, we optimize over it,  allowing the $\theta$ parameters to change slightly since the given pose $\theta_{\text{init}}$ might not be well supported by the scene. 
This allows for small pose adjustment that might be necessary to better fit the body into the scene. $\lambda_{reg} = 100$.

The input posed mesh, $\meshBody$, can come from any source.
For example, we can generate random \smplX meshes using \vposer~ \cite{SMPL-X:2019} which is a VAE trained on a large dataset of human poses. 
More interestingly, we use \smplX meshes fit to realistic \renderpeople scans \cite{agora2020} (see Fig.~\ref{fig:teaser}). 

\newcommand{\affordDetectHeight}{0.495}
\begin{figure}
	\centering
	\includegraphics[trim=000mm 000mm 000mm 000mm, clip=true, width=0.96 \linewidth]{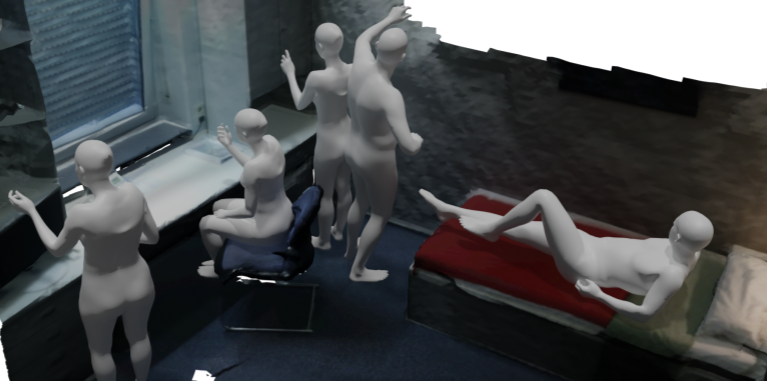}
	\includegraphics[trim=000mm 000mm 000mm 000mm, clip=true, width=0.96 \linewidth]{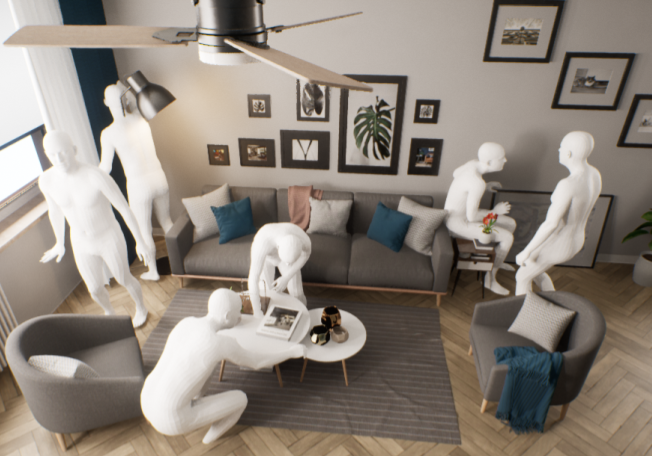}
\vspace{-0.1in}
	\caption{
				\textbf{(Top):} 
				\smplX meshes automatically placed in a real scene from the \prox  test set. 
				The body shapes and poses here are drawn from the \prox test set and were not used in training.
				\textbf{(Bottom):} 
				\smplX meshes automatically placed in a synthetic scene.
	}
	\label{fig:afford}
\end{figure}

We tested our method with both real (scanned) and synthetic (artist generated) scenes. 
Example bodies optimized to fit in a real scene from the \prox \cite{PROX:2019} test set are shown in Fig. \ref{fig:afford} (top); this scene was not used during training.
Note that people appear to be interacting naturally with the scene; that is, their pose matches the scene context.
Figure \ref{fig:afford} (bottom) shows bodies automatically placed in an artist-designed scene (Archviz Interior Rendering Sample, Epic Games)\footnote{\url{https://docs.unrealengine.com/en-US/Resources/Showcases/ArchVisInterior/index.html}}.
\modelname goes beyond previous work \cite{grabner2011chair,shape2pose2014,zhang2020generating} to produce realistic human-scene interactions for a wide range of poses like lying down and reaching out.

While the poses look natural in the above results, the \smplX bodies look out of place in realistic scenes.
Consequently, we would like to render realistic people instead, but models like \smplX do not support realistic clothing and textures. 
In contrast, scans from companies like \renderpeople (Renderpeople GmbH, K\"{o}ln) are realistic, but have a different mesh topology for every scan.
The consistent topology of a mesh like \smplX is critical to learn the feature model. 

\textbf{Clothed Humans:} 
We address this issue by using \smplX fits to clothed meshes from the \agora dataset \cite{agora2020}. 
We then take the \smplX fits and minimize an energy function similar to Eq.~\ref{eq:afford} with one important change. 
We keep the pose, $\theta$, fixed:
\begin{equation}
E(\trans, \theta_0) =  \mathcal{L}_{\mathit{afford}} + \mathcal{L}_{\mathit{pen}} . \label{eq:afford_cloth}
\end{equation}
Since the pose does not change, we just replace the \smplX mesh with the original clothed mesh after the optimization converges; 
see \supmat for details.

\begin{figure}[t]
	\centering
	\includegraphics[trim=000mm 000mm 000mm 000mm, clip=true, width=1.00 \linewidth]{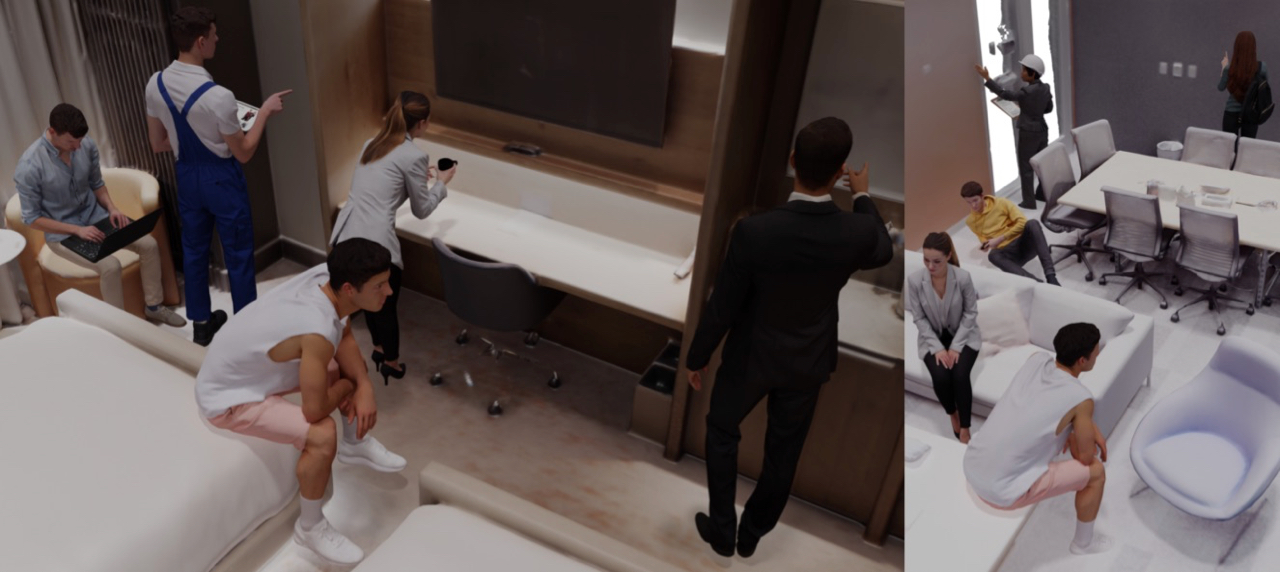}	
\vspace{-0.24in}
	\caption{Unmodified clothed bodies (from \renderpeople) automatically placed in real scenes from the Replica dataset.}
	\label{fig:affordance_replica_rp}
\end{figure}

Qualitative results for real scenes (Replica dataset \cite{replica19arxiv}) are shown in Fig.~\ref{fig:affordance_replica_rp}, and for a synthetic scene in Fig.~\ref{fig:teaser}. 
More results are shown in \supmat and in our \video. 

\subsubsection{Evaluation}						
\label{sec:eval_PerceptualStudy}		We quantiatively evaluate \modelname with
two perceptual studies. 
In both, %
subjects are shown a pair of two rendered scenes, and must choose the one that best answers the question
``Which one of the two examples has more realistic (i.e. natural and physically plausible) human-scene interaction?''
We also evaluate physical plausibility and diversity.

\textbf{Comparison to \prox ground truth:}
We follow the protocol of Zhang \etal \cite{PLACE:3DV:2020} and compare our results to randomly selected examples from \prox ground truth. 
We take $4$ real scenes from the \prox \cite{PROX:2019} test set, namely MPH16, MPH1Library, N0SittingBooth and N3OpenArea. 
We take $100$ \smplX bodies from the \agora \cite{agora2020} dataset, corresponding to $100$ different \threeD scans from \renderpeople.
We take each of these bodies and sample one \featMap for each, using our cVAE.
We then automatically optimize the placement of each sample in all the scenes, one body per scene.
For unclothed bodies (Tab.~\ref{table:affordanceEval___GENERATION_vs_PROX_GT}, rows 1-3), this optimization changes the pose slightly to fit the scene (Eq.~\ref{eq:afford}).
For clothed   bodies (Tab.~\ref{table:affordanceEval___GENERATION_vs_PROX_GT}, rows 4-5), the pose is kept fixed (Eq.~\ref{eq:afford_cloth}).
For each variant, this optimization results in $400$ unique body-scene pairs.
We render each \threeD human-scene interaction from $2$ views so that subjects are able to get a good sense of the \threeD relationships from the images.
Using Amazon Mechanical Turk (AMT), we show these results to $3$ different subjects.
This results in $1200$ unique ratings.
The results are shown in Tab.~\ref{table:affordanceEval___GENERATION_vs_PROX_GT}. 
\modelname (contact only) and the state-of-the-art method \place \cite{PLACE:3DV:2020} are both almost indistinguishable from the \prox ground truth.
However, the proposed \modelname (contact + semantics) (row $3$) outperforms both \modelname (contact only) (row $2$) and \place \cite{PLACE:3DV:2020} (row $1$), thus modeling scene semantics increases realism.
Lastly, the rendering of high quality clothed meshes (bottom two rows) influences the perceived realism significantly. 

\begin{table}[]
	\centering
	\small
	\begin{tabular}{lcc}
	\hline
	 \hline
		{}										& 	\multicolumn{1}{r}{Generation $\uparrow$}	& \prox GT $\downarrow$	\\ \hline \hline
		\place \cite{PLACE:3DV:2020}				& 	$48.5\%$  									& $51.5\%$				\\ \hline
		\modelname (contact only) 				& 	$46.9\%$  									& $53.1\%$				\\
		\modelname (contact + semantics)			& 	$\boldsymbol{49.1\%}$  						& $50.1\%$				\\ \hline
		\modelname-\emph{clothing} (contact)		& 	$55.0\%$  									& $45.0\%$				\\
		\modelname-\emph{clothing} (semantics)	& 	$\boldsymbol{60.6\%}$  						& $39.4\%$				\\
		\hline
	\end{tabular}
\vspace{-0.1in}
	\caption{
				Comparison to \prox \cite{PROX:2019}
                                ground truth. %
				Subjects are shown pairs of a generated
                                \threeD human-scene interaction and
                                \prox ground truth (GT), and must
                                chose the most realistic one. 
                                A higher percentage means that
                                subjects deemed this method more realistic.
	}
	\label{table:affordanceEval___GENERATION_vs_PROX_GT}
\end{table}

\begin{table}[]
	\centering
	\small
	\begin{tabular}{lcc}
	\hline
	 \hline
		{}										&	\multicolumn{1}{r}{\modelname-variant $\uparrow$}		& \place $\downarrow$	\\ \hline  \hline
		\modelname (contact only)				& 	$60.7\%$  											& $39.3\%$				\\
		\modelname (contact + semantics) 		& 	$\boldsymbol{61.0\%}$  								& $39.0\%$				\\
		\hline
	\end{tabular}
\vspace{-0.1in}
	\caption{
				\modelname compared to \place for
                                \threeD human-scene interaction
                                generation. %
See Tab.~\ref{table:affordanceEval___GENERATION_vs_PROX_GT} caption.
	}
	\label{table:affordanceEval___POSA_vs_PLACE}
\end{table}

\textbf{Comparison between \posa and \place:}
We follow the same protocol as above, but this time we directly compare \posa and \place. 
The results are shown in Tab.~\ref{table:affordanceEval___POSA_vs_PLACE}.
Again, we find that adding semantics improves realism.
There are likely several reasons that \posa is judged more realistic than \place.
First,  \posa employs denser \jargonContact information across the whole \smplX body surface, compared to \place's sparse distance information through its \bps representation. 
Second, \posa uses a human-centric formulation, as opposed to \place's scene-centric one, and this can help generalize across scenes better. 
Third,  \posa uses semantic features that help bodies do the right thing in the scene, while \place does not. 
When human generation is imperfect, inappropriate semantics may make the result seem worse. 
Fourth, the two methods are solving slightly different tasks. 
\place generates a posed body mesh for a given scene, while our method samples one from the \agora dataset and places it in the scene using a generated \modelname~ \featMap. 
While this gives \place an advantage, because it can generate an appropriate pose for the scene, it also means that it could generate an unnatural pose, hurting realism.
In our case, the poses are always ``valid'' by construction but may not be appropriate for the scene.
Note that, while more realistic than prior work, the results are not always fully natural; sometimes people sit in strange places or lie where they usually would not.

\label{sec:eval_PhysicalDiversity}	
\textbf{Physical Plausibility}:
We take $1200$ bodies from the \agora~\cite{agora2020} dataset and place all of them in each of the $4$ test scenes of \prox, leading to a total of $4800$ samples, following \cite{PLACE:3DV:2020,zhang2020generating}. 
Given a generated body mesh, $\meshBody$, a scene mesh, $\meshScene$, and a scene signed distance field (SDF) that stores distances $d_j$ for each voxel $j$, we compute the following scores, defined by Zhang \etal~\cite{zhang2020generating}: 
(1) the \emph{non-collision score} for each $\meshBody$, which is the ratio of body mesh vertices with positive SDF values divided by the total number of \smplX vertices, and 
(2) the \emph{contact score}       for each $\meshBody$, which is $1$ if at least one vertex of $\meshBody$ has a non-positive value. 
We report the mean non-collision score and mean contact score over all $4800$ samples in Tab. \ref{table:affordanceEval_physical}; higher values are better for both metrics. 
\posa and \place are comparable under these metrics.

\textbf{Diversity Metric}:
Using the same $4800$ samples, we compute the diversity metric from \cite{zhang2020generating}.
We perform K-means ($k=20$) clustering of the \smplX parameters of all sampled poses, and report: 
(1) the entropy of the cluster sizes, and 
(2) the cluster size, \ie~the average distance between the cluster center and the samples belonging in it. 
See Tab.~\ref{table:affordanceEval_diversity}; higher values are better. 
While \place generates poses and \posa samples them from a database, there is little difference in diversity.

\begin{table}[]
	\centering
	\small
	\begin{tabular}{lcc}
	\hline
	 \hline
		&\multicolumn{1}{r}{Non-Collision $\uparrow$}	& Contact $\uparrow$							\\ \hline \hline
		PSI \cite{zhang2020generating}					& $0.94$  				& $0.99$				\\ \hline
		\place \cite{PLACE:3DV:2020}						& $\boldsymbol{0.98}$  	& $0.99$				\\ \hline
		\modelname (contact only) 						& $0.97$  				& $\boldsymbol{1.0}$	\\
		\modelname (contact + semantics)					& $0.97$  				& $0.99$				\\ \hline
		\hline
	\end{tabular}
\vspace{-0.1in}
	\caption{
				Evaluation of the physical plausibility metric. 
				Arrows indicate that higher scores are better. 
	}
	\label{table:affordanceEval_physical}
\end{table}

\begin{table}[]
	\centering
	\small
	\begin{tabular}{lcc}
		\hline
		\hline
		&\multicolumn{1}{r}{Entropy $\uparrow$}							& Cluster Size $\uparrow$	\\ \hline \hline
		PSI \cite{zhang2020generating}			& $\boldsymbol{2.97}$ 	& $2.53$						\\ \hline
		\place \cite{PLACE:3DV:2020}				& $2.91$  				& $\boldsymbol{2.72}$		\\ \hline
		\modelname (contact only) 				& $2.94$  				& $2.28$						\\
		\modelname (contact + semantics)			& $2.92$  				& $2.27$						\\ \hline
		\hline
	\end{tabular}
\vspace{-0.1in}
	\caption{
				Evaluation of the diversity metric. 
				Arrows indicate that higher scores are better. 
	}
	\label{table:affordanceEval_diversity}
\end{table}

\subsection{Monocular Pose Estimation with HSI}				\label{sec:pose_estimation}			Traditionally, monocular pose estimation methods focus only on the body and ignore the scene. 
Hence, they tend to generate bodies that are inconsistent with the scene. 
Here, we compare directly with  \prox \cite{PROX:2019}, which adds contact and penetration constraints to the pose estimation formulation. 
The contact constraint snaps a fixed set of contact points on the body surface to the scene, if they are ``close enough''. 
In \prox, however, these contact points are manually selected and are independent of pose.

We replace the hand-crafted contact points of \prox with our learned \featMap.
We fit \smplX to \rgb image features such that the contacts are consistent  with the \threeD scene and its semantics.
Similar to \prox, we build on \smplifyX \cite{SMPL-X:2019}. %
Specifically, \smplifyX optimizes \smplX parameters to minimize an objective function of multiple terms: 
the re-projection error of \twoD joints, priors and physical constraints on the body; 
$E_\text{\smplifyX}(\beta,\theta,\psi,\trans) =$
\begin{equation}
					E_J 			+
\lambda_{\theta} 	E_{\theta}	+
\lambda_{\alpha}	 	E_{\alpha}	+ 
\lambda_{\beta}    	E_{\beta}	+ 
\lambda_{\mColl}    	E_{\mColl}
\label{eq:smplify_objective}
\end{equation}
where $\theta$ represents the pose parameters of the body, face (neck, jaw) and the two hands, $\theta=\{ \theta_\body, \theta_\face, \theta_\hand \}$, $\trans$ denotes the body translation, and $\beta$ the body shape.
$E_J$ is a re-projection loss that minimizes the difference between \twoD joints estimated from the \rgb image $\imgRGB$ and the \twoD projection of the corresponding posed \threeD joints of \smplX.
$E_{\alpha}(\theta_\body) = \sum_{i \in (elbows,knees)}\exp(\theta_i)$  is a prior penalizing extreme bending only for elbows and knees.
The term $E_{\mColl}$ penalizes self-penetrations. For details please see \cite{SMPL-X:2019}.

We turn off the \prox contact term and
optimize Eq.~\ref{eq:smplify_objective} to get a pose matching the image observations and roughly obeying  scene constraints.
Given this rough body pose, which is not expected to change significantly, we sample features from $P(\featGen|z,\bodyV)$ and keep these fixed.
Finally, we refine by minimizing  $E(\beta,\theta,\psi,\trans,\meshScene) = $
\begin{equation}
E_\text{\smplifyX} +  ||{\featGen}_{c} \cdot \feat_d|| + \mathcal{L}_{pen} 
\end{equation}
where 
$E_\text{\smplifyX}$ 	represents the \smplifyX energy term as defined in Eq.~\ref{eq:smplify_objective},
${\featGen}_{c}$ 		are the generated contact labels, 
$\feat_d$ 				is the observed distance, and $\mathcal{L}_{pen}$ represents the body-scene penetration loss as in Eq.~\ref{eg:penetration}. 
We compare our results to standard \prox in Tab.~\ref{table:pose_estimation}. 
We also show the results of \rgb\/-only baseline introduced in \prox for reference. 
Using our learned feature map improves accuracy over the \prox\/'s heuristically determined contact constraints.

\begin{table}[]
	\centering
	\small
	\begin{tabular}{lllll}
	\hline
	\hline
		(mm)			& PJE $\downarrow$      			& V2V $\downarrow$ 				& p.PJE  $\downarrow$ 			& p.V2V $\downarrow$ 			\\ 	\hline
		RGB 			& $220.27$ 			& $218.06$ 			& $73.24$ 			& $60.80$ 			\\ 
		\prox 		& $167.08$ 			& $166.51$ 			& $\mathbf{71.97}$ 	& $\mathbf{61.14}$ 			\\ 
		\modelname 	& $\mathbf{154.33}$ 	& $\mathbf{154.84}$ 	& $73.17$			& $63.23$ 	\\ 
		\hline
	\end{tabular}
\vspace{-0.1in}
	\caption{
				Pose estimation results for \prox and \modelname.
				\emph{PJE} 			is the mean per-joint error and 
				\emph{V2V} 			is the mean vertex-to-vertex Euclidean distance between meshes (after only pelvis joint alignment). 
				The prefix ``p'' 	means that the
                                error is computed  after Procrustes
                                alignment to the ground truth; this hides many errors, making the methods comparable.
	}
	\label{table:pose_estimation}
\end{table}

\section{Conclusions}

Traditional \threeD body models, like \smplX, model the a priori probability of possible body shapes and poses.
We argue that human poses in isolation from the scene, make little sense. 
We introduce \modelname, which effectively upgrades a \threeD human body model to explicitly represent possible human-scene interactions.
Our novel, body-centric, representation encodes the \jargonContact and semantic relationships between the body and the scene.
We show that this is useful and supports new tasks.
For example, we consider placing a \threeD human into a \threeD scene.  
Given a scan of a person with a known pose, \modelname allows us to search the scene for locations where the pose is likely.
This enables us to populate empty \threeD scenes with higher realism than the state of the art.
We also show that \modelname can be used for estimating human pose
from an \rgb image,
and that the body-centered HSI representation improves accuracy.
In summary, \modelname is a good step towards a richer model of human bodies that goes beyond pose to support the modeling of \hsi.

\noindent
\textbf{Limitations:}
\posa requires an accurate scene SDF;
a noisy scene mesh can lead to penetration between the body and scene.
\modelname focuses on a single body mesh only.  
Penetration between clothing and the scene is not handled and multiple
bodies are not considered.
Optimizing the placement of people in scenes is sensitive to initialization and is prone to local minima. 
A simple user interface would address this, letting naive users
roughly place bodies, and then \posa would automatically refine the placement.

{
\noindent
\small
\textbf{Acknowledgements:}
We thank M.~Landry for the video voice-over, B.~Pellkofer for the project website, and R.~Danecek and H.~Yi for proofreading. 
This work was partially supported by the International Max Planck Research School for Intelligent Systems (\mbox{IMPRS-IS}).
\textbf{Disclosure:}
MJB has received research funds from Adobe, Intel, Nvidia, Facebook, and Amazon. While MJB is a part-time employee of Amazon, his research was performed solely at, and funded solely by, Max Planck. MJB has financial interests in Amazon, Datagen Technologies, and Meshcapade GmbH.
}

{
\small
\balance
\bibliographystyle{ieee_fullname}
\bibliography{arxiv_UPLOAD_posa_CLEAN_BIB}
}

\renewcommand{\thefigure}{S.\arabic{figure}}
\setcounter{figure}{0}
\renewcommand{\thetable}{S.\arabic{table}}
\setcounter{table}{0}

\newpage
\begin{appendices}
	\begin{figure*}
	\centering	
	\includegraphics[trim=000mm 000mm 000mm 000mm, clip=false, width=1.00 \textwidth]{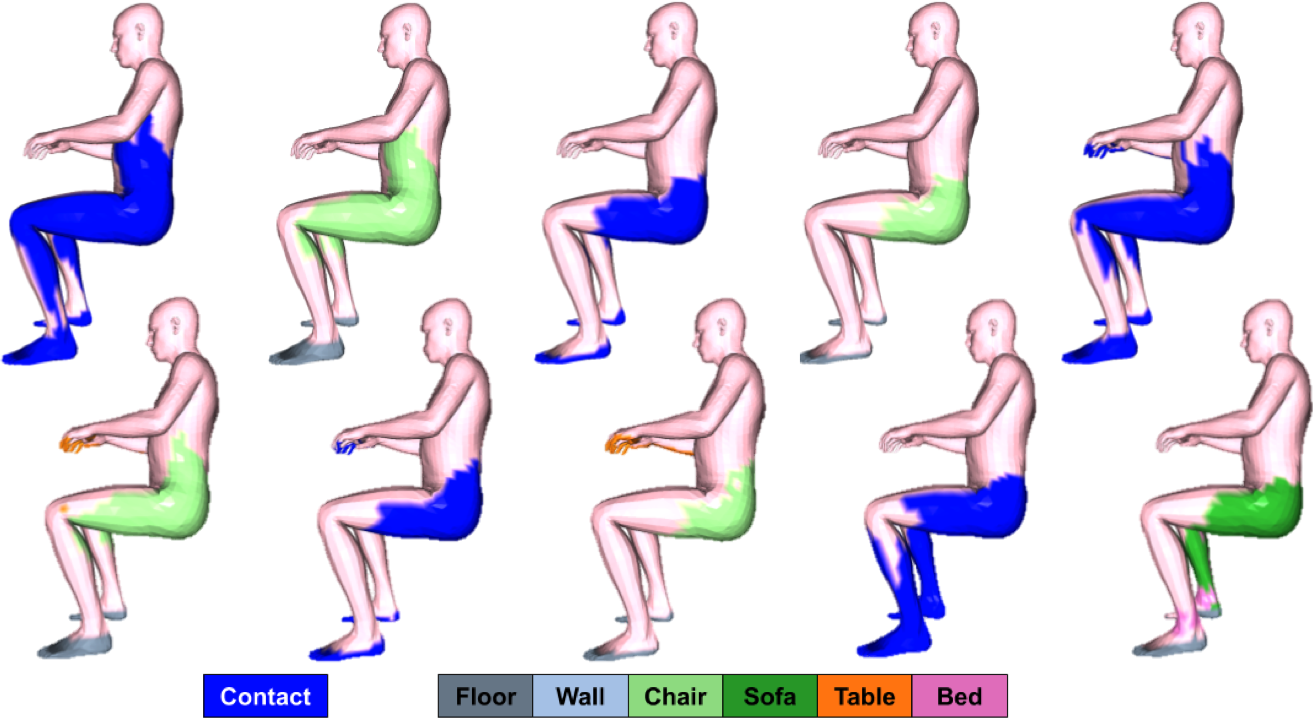}
	\caption{
		Random samples from our trained cVAE for the same pose. 
		For each example we show from left to right: $\featCon$ and $\featSem$.
		The color code is at the bottom.  
		For $\featCon$, blue means contact, while pink means no contact.  
		For $\featSem$, each scene category has a different color.
    }
    \label{fig:SupMat_random_samples}
\end{figure*}

	\section{Training Details}

The global orientation of the body is typically irrelevant in our body-centric representation, so we rotate the training bodies around the $y$ and $z$ axes to put them in a canonical orientation.
The rotation around the $x$ axis, however, is essential to enable the model to differentiate between standing up and lying down. 
The semantic labels for the  \prox scenes are taken from Zhang \etal \cite{zhang2020generating}, where scenes were manually labeled following the object categorization of Matterport3D \cite{Matterport3D}, which incorporates $40$ object categories.  

Our encoder-decoder architecture is similar to the one introduced in Gong \etal \cite{gong2019spiralnet++}. 
The encoder consists of $3$ spiral convolution layers interleaved with pooling layers $3 \times \left\{ \text{Conv} \left( 64\right) \rightarrow \text{Pool} \left(4\right) \right\} \rightarrow \text{FC} \left(512\right) $. 
Pool stands for a downsampling operation as in COMA \cite{Ranjan_2018_ECCV}, which is based on contracting vertices. 
FC is a fully connected layer and the number in the bracket next to it denotes the number of units in that layer.
We add $2$ additional fully connected layers to predict the parameters of the latent code, with fully connected layers of $256$ units each. 
The input to the encoder is a body mesh $\meshBody$ where, for each vertex, $i$, we concatenate $\queryBodyV$ vertex positions, and $\feat$ vertex features. 
For computational efficiency, we first downsample the input mesh by a factor of $4$. 
So instead of working on the full mesh resolution of $10475$ vertices, our input mesh has a resolution of $655$ vertices. 
The decoder architecture consists of spiral convolution layers only $ 4 \times \left\{ \text{Conv} \left( 64\right)\right\} \rightarrow  \text{Conv} \left(N_f\right)$. 
We attach the latent vector $z$ to the \threeD coordinates of each vertex similar to Kolotouros \etal \cite{kolotouros2019cmr}.

We build our model using the \pytorch framework. 
We use the Adam optimizer \cite{kingma2014adam}, batch size of $64$, and learning rate of $1e^{-3}$ without learning rate decay.

	\section{SDF Computation}

For computational efficiency, we employ a precomputed \threeD signed distance field (SDF) for the static scene $\surfaceScene$, following Hassan \etal \cite{PROX:2019}. 
The SDF has a resolution of $512 \times 512 \times 512$. 
Each voxel $c_j$ stores the distance $d_j \in \mathbb{R}$ of its centroid $P_j \in \mathbb{R}^3$ to the nearest surface point $\closestSceneP \in \surfaceScene$. 
The distance $d_j$ has a positive sign if $P_j$ lies in the free space outside physical scene objects, while it has a negative sign if it is inside a scene object. 

	\section{Random Samples}
We show multiple randomly sampled \featMaps for the same pose in Fig.~\ref{fig:SupMat_random_samples}. 
Note how \posa generate a variety of valid \featMaps for the same pose. 
Notice for example that the feet are always correctly predicted to be in contact with the floor. 
Sometimes our model predicts the person is sitting on a chair (far left) or on a sofa (far right). 

The predicted semantic map $\featSem$ is not always accurate as shown in the far right of Fig.~\ref{fig:SupMat_random_samples}. 
The model predicts the person to be sitting on a sofa but at the same time predicts the lower parts of the leg to be in contact with a bed which is unlikely. 

	\begin{figure*}[t]
	\centering
	\includegraphics[width=1.00 \linewidth]{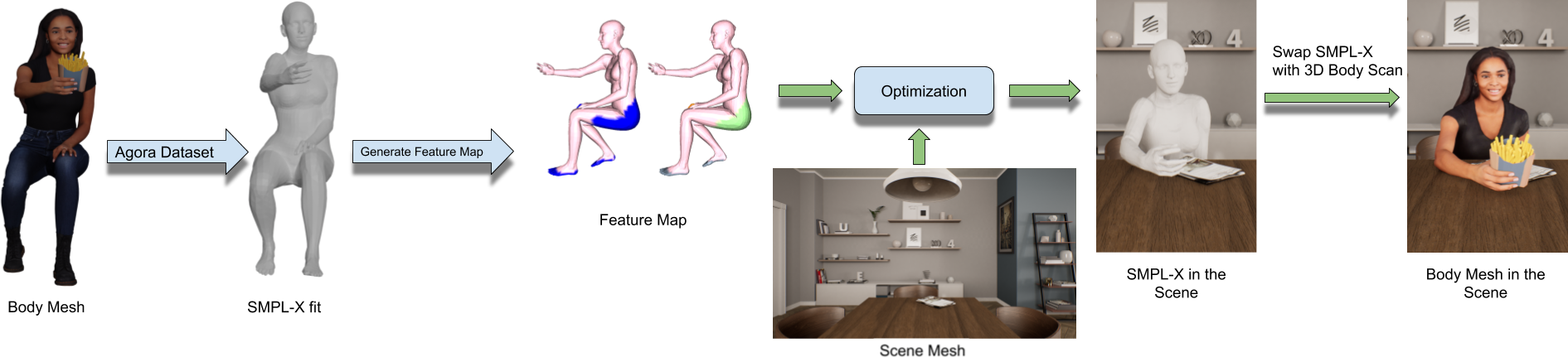}	
	\vspace{-2.0em}
	\caption{
		Putting realistic people in scenes. 
		Pipeline of affordance detection using meshes with clothing. 
		\smplX acts as a proxy for the clothed scan. 
		\modelname is used to sample features for this pose. 
		These features are then  used with the scene mesh to optimize the placement of the body. 
		After convergence, we simply replace \smplX with the clothed scan. 
	}
	\label{fig:SupMat_pipeline}
\end{figure*}

\section{Affordance Detection}

The complete pipeline of the affordance detection task is shown in Fig.~\ref{fig:SupMat_pipeline}. 
Given a clothed \threeD mesh that we want to put in a scene, we first need a \smplX fit to the mesh; here we take this from the \agora dataset \cite{agora2020}. 
Then we generate a \featMap using the decoder of our cVAE by sampling $P(\featGen|z,\bodyV)$. 
Next we minimize the energy function in Eq.~\ref{eq:SupMat_afford_cloth}. 
\begin{equation}
E(\trans, \theta_0) =  \mathcal{L}_{\mathit{afford}} + \mathcal{L}_{\mathit{pen}}  \label{eq:SupMat_afford_cloth}
\end{equation} 
Finally, we replace the \smplX mesh with the original clothed.

\begin{figure*}
	\centering
	\includegraphics[width=0.90 \linewidth]{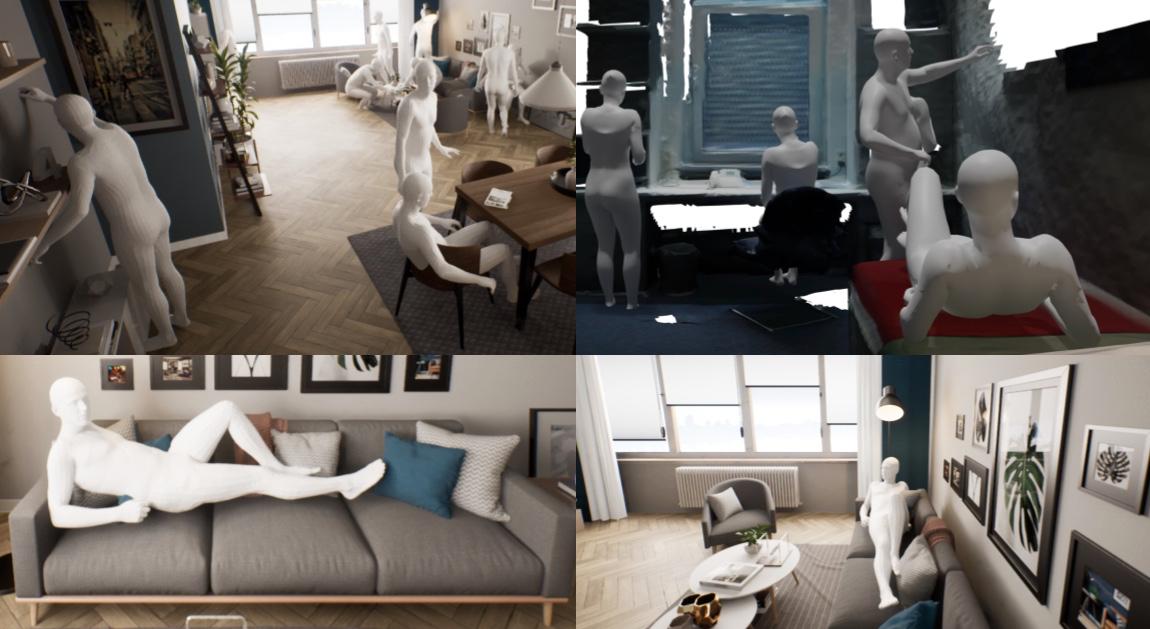}	
	\caption{
				Qualitative examples of \smplX meshes automatically placed in real and synthetic scenes. 
				The body shapes and poses were not used in training.
	}
	\label{fig:SupMat_smplx_qualitatives}
\end{figure*}

\begin{figure*}
	\centering
	\includegraphics[width=.90 \linewidth]{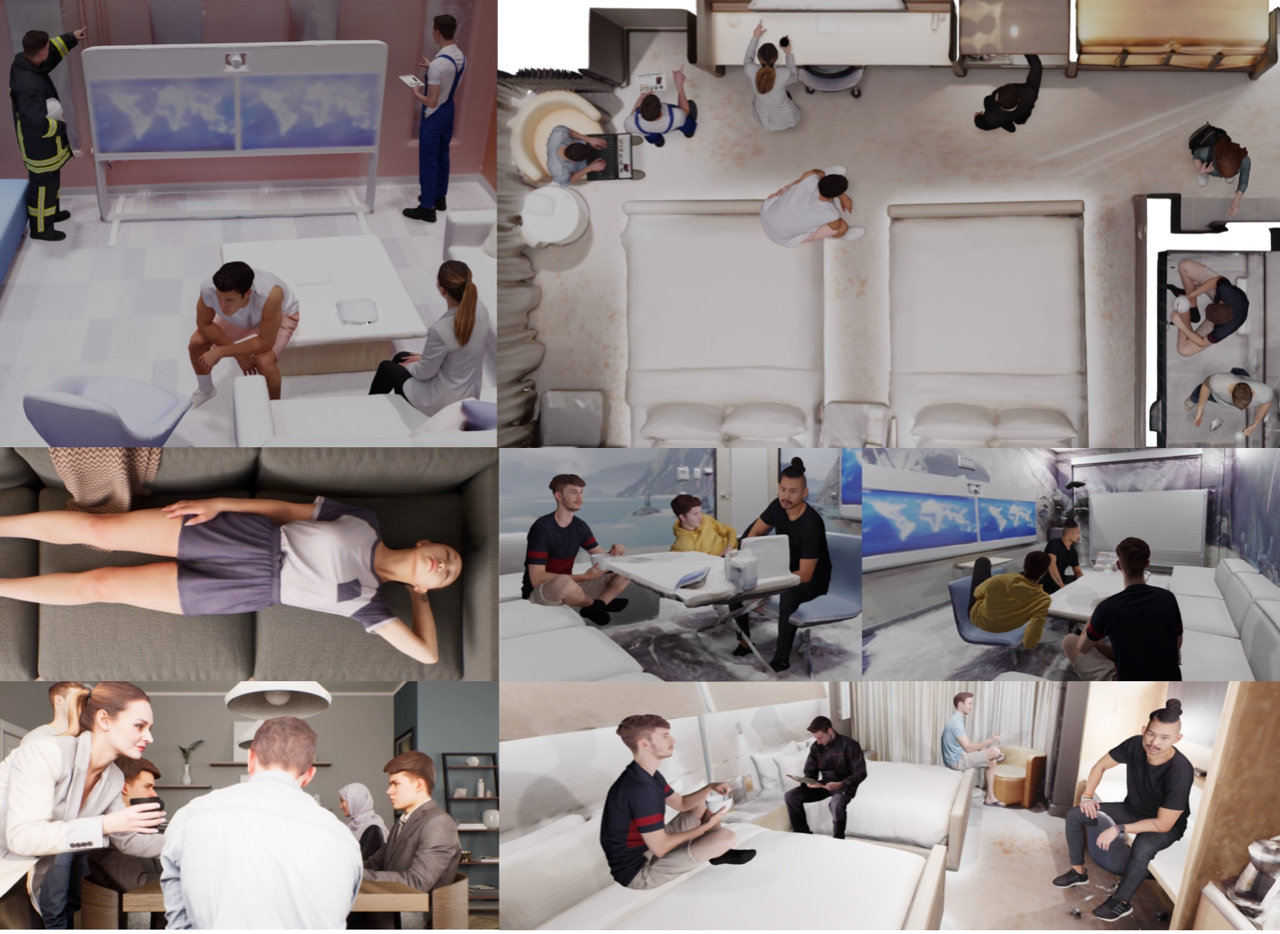}	
	\caption{Clothed bodies (from \renderpeople) automatically placed in real and synthetic scenes.}
	\label{fig:SupMat_rp_qualitatives}
\end{figure*}

\begin{figure*}
	\centering
	\includegraphics[width=.90 \linewidth]{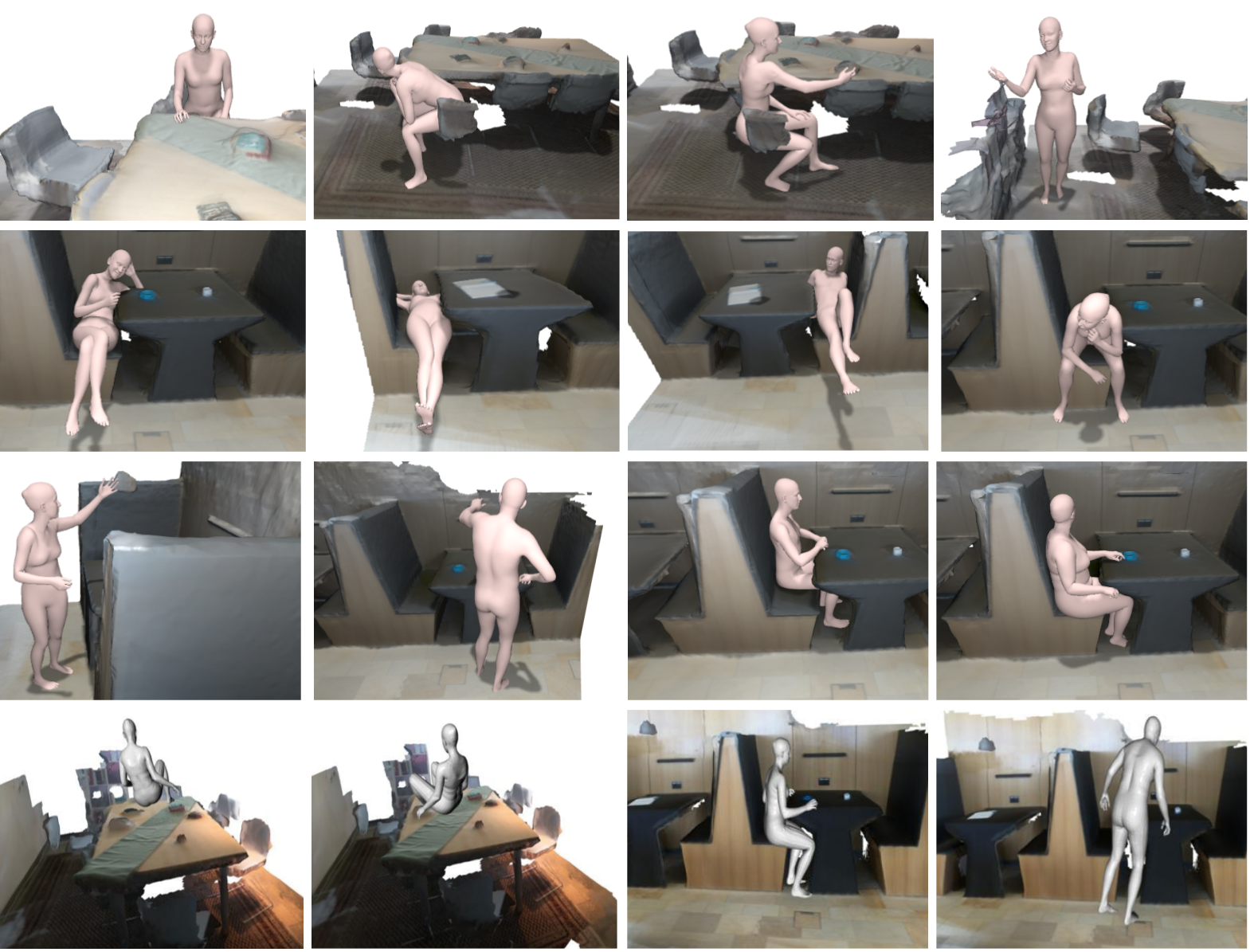}	
	\caption{Qualitative examples from \modelname (pink) and PLACE~\cite{PLACE:3DV:2020} (silver).}
	\label{fig:SupMat_posa_place}
\end{figure*}

We show additional qualitative examples of \smplX meshes automatically placed in real and synthetic scenes in Fig.~\ref{fig:SupMat_smplx_qualitatives}. 
Qualitative examples of clothed bodies placed in real and synthetic scenes are shown in Fig.~\ref{fig:SupMat_rp_qualitatives}. We show qualitative comparison between our results and PLACE~\cite{PLACE:3DV:2020} in Fig.~\ref{fig:SupMat_posa_place}.

	\section{Failure Cases}

\begin{figure*}[t]
	\centering
	\includegraphics[trim=000mm 000mm 000mm 000mm, clip=true, width=0.98 \linewidth]{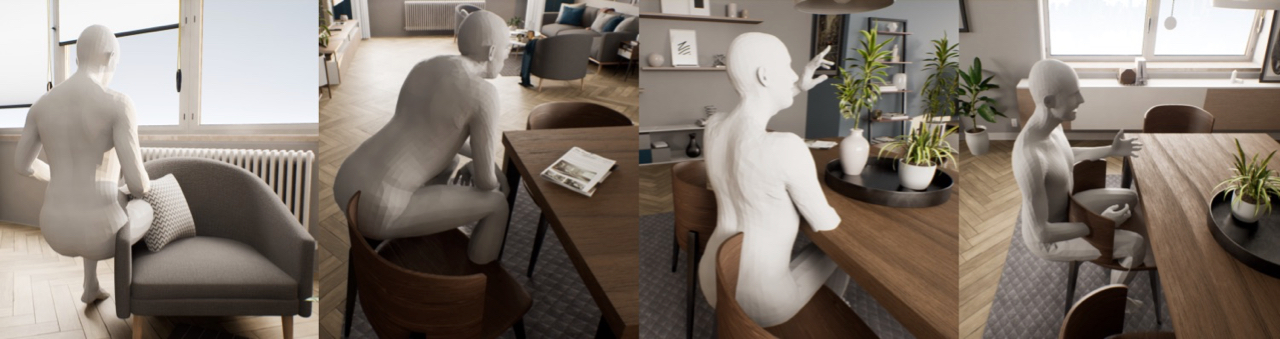}	
	\vspace{-0.5em}
	\caption{Failure cases.}
	\label{fig:SupMat_failure_cases}
\end{figure*}

We show representative failure cases in Fig. \ref{fig:SupMat_failure_cases}. 
A common failure mode is residual penetrations; even with the penetration penalty the body can still penetrate the scene. 
This can happen due to thin surfaces that are not captured by our SDF and/or because the optimization becomes stuck in a local minimum. 
In other cases, the feature map might not be right. 
This can happen when the model does not generalize well to test poses due to the limited training data.

	\section{Effect of Shape}
\begin{figure*}
	\centering
	\includegraphics[width= \linewidth]{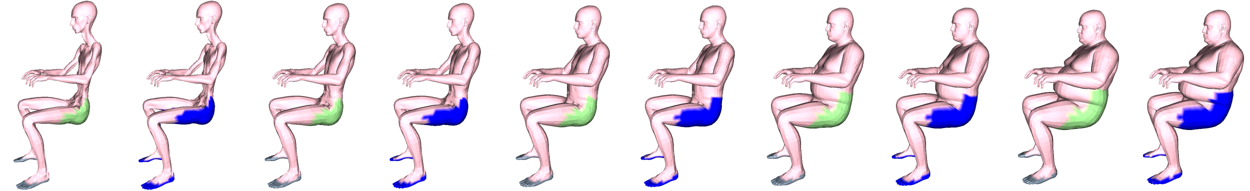}
	\vspace{-1.1em}
	\caption{Generated \featMaps for various body shapes.}
	\label{fig:SupMat_betas}
	\vspace{-0.5em}
\end{figure*}

Fig.~\ref{fig:SupMat_betas} shows that our model can predict plausible  feature maps for a wide range of human body shapes. 

	\section{Scene population.}
\begin{figure*}
	\includegraphics[trim=000mm 000mm 005mm 000mm, clip=true, width=1.05 \linewidth]{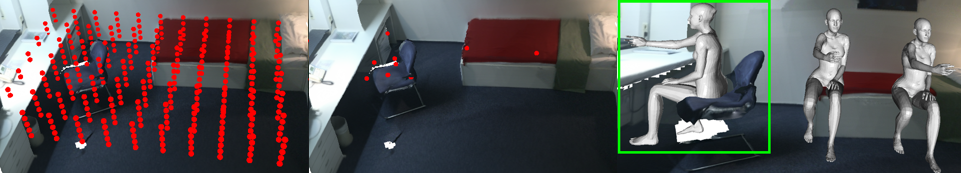}%
	\caption{
		Main steps of our method for scene population. 
		\textbf{(1)} Grid with all candidate positions. 
		\textbf{(2)} The $10$ best positions. 
		\textbf{(3)} Final result.
	}
	\label{fig:auto_placement}
\end{figure*}

In Fig.~\ref{fig:auto_placement} we show the three main steps to populate a scene:
\textbf{(1)}		Given a scene, we create a regular grid of candidate positions (Fig.~\ref{fig:auto_placement}\redify{.1}). 
				We place the body, in a given pose, at each candidate position and evaluate \mbox{Eq. \redify{10}} once. 
\textbf{(2)}		We then keep the $10$ best candidates with the lowest energy (Fig.~\ref{fig:auto_placement}\redify{.2}), 
				and 
\textbf{(3)}		iteratively optimize \mbox{Eq.~\redify{10}} for these; %
				Fig.~\ref{fig:auto_placement}\redify{.3} shows 
				results at three positions, with the best one highlighted with green.

\end{appendices}

\end{document}